\documentclass[]{academic_template}

\usepackage[utf8]{inputenc}
\usepackage[T1]{fontenc}
\usepackage{lmodern}
\usepackage{url}
\usepackage{amsfonts}
\usepackage{amsmath}
\usepackage{amssymb}
\usepackage{amsthm}

\usepackage{nicefrac}
\usepackage{colortbl}
\usepackage{array}
\usepackage{enumitem}
\usepackage{wrapfig}
\usepackage{makecell}
\usepackage[most]{tcolorbox}
\definecolor{LighterGray}{RGB}{248,248,250}
\definecolor{DeepPurple}{RGB}{69,45,117}
\definecolor{White}{RGB}{255,255,255}
\definecolor{CaseOrange}{RGB}{184,83,24}
\definecolor{CaseBlue}{RGB}{38,91,166}

\newcommand{\igbenchfull}{IdeaGene-Bench}
\newcommand{\igbench}{IG-Bench}
\newcommand{\gene}{\igbench}
\newcommand{\geneexam}{IG-Exam}
\newcommand{\genearena}{IG-Arena}
\newcommand{\ideagene}{IdeaGene}
\newcommand{\genome}{\textit{Idea Genome}}
\newcommand{\genomediff}{GenomeDiff}
\newcommand{\yesmark}{\ensuremath{\textcolor{green!55!black}{\checkmark}}}
\newcommand{\nomark}{\ensuremath{\textcolor{red!70!black}{\times}}}

\setlogoheight{14mm}
\setlogospacing{5mm}
\setsjtublue
\settitlerulethickness{3pt}
\abstractboxon
\setabstractframecolor{gray}
\setabstractbgcolor{gray!10}
\setlogotolineshift{5mm}
\settoprulethickness{2.5pt}
\setbottomrulethickness{1.5pt}

\title{Ideas Have Genomes:\\
Benchmarking Scientific Lineage Reasoning\\
and Lineage-Grounded Idea Generation}

\author[1,*]{Yifan Zhou}
\author[1,*]{Qihao Yang}
\author[1,*]{Yan Li}
\author[1,*]{Donggang Li}
\author[1]{Xiru Hu}
\author[2]{Hokin Deng}
\author[1]{Ziyang Gong}
\author[1]{Xuanyi Zhou}
\author[3]{Huacan Wang}
\author[4]{Xiangchao Yan}
\author[1]{Wanghan Xu}
\author[4]{Wenlong Zhang}
\author[5]{Shaofeng Zhang}
\author[6]{Yue Zhou}
\author[7]{Yifan Yang}
\author[1]{Zhihang Zhong}
\author[1,\dagger]{Xue Yang}

\affiliation[1]{Shanghai Jiao Tong University}
\affiliation[2]{Carnegie Mellon University}
\affiliation[3]{University of Chinese Academy of Sciences}
\affiliation[4]{Shanghai Artificial Intelligence Laboratory}
\affiliation[5]{University of Science and Technology of China}
\affiliation[6]{East China Normal University}
\affiliation[7]{Microsoft Corporation}

\contribution[*]{Equal contribution}
\contribution[\dagger]{Corresponding Author}

\abstract{
Scientific ideas rarely start from a blank page. They inherit mechanisms, repair known limitations, and recombine pieces of earlier work, much like biological genomes. Current benchmarks still say little about whether AI systems can follow this inheritance structure. We present \textbf{\igbenchfull{} (\igbench{})}, a benchmark for scientific lineage reasoning and lineage-grounded idea generation. \igbench{} is organized around the \ideagene{} framework: each paper or proposal is represented as a set of minimal, typed, evidence-grounded \genome{} objects, and a \genomediff{} aligns these objects to record inheritance, mutation, loss, external import, and novel insertion under six operational evolutionary dynamics. The benchmark contains 1,961 golden lineage traces, 1,085 curated \genome{} objects, and 920 pairwise \genomediff{} records across 10 scientific domains. It supports two evaluations. \geneexam{} (42 task types, 1,029 instances) tests closed-form lineage reasoning across \genome{} abstraction, inheritance tracing, evolutionary reasoning, and lineage verification. \genearena{} evaluates generation with a lineage-conditioned Population-Evolution Score (PES), asking whether a proposal can be inserted as a coherent descendant of a given lineage population: it should inherit the right \genome{} objects, vary meaningfully from nearby work, and offer selection value for future research. Experiments on 14 LLM-based scientists expose a compositional bottleneck. The strongest system reaches only 27.3\% exact accuracy on lineage reasoning, and structured lineage context reshuffles system rankings rather than helping every participant uniformly.
}

\date{\today}
\correspondence{\email{yangxue-2019-sjtu@sjtu.edu.cn}}
\checkdata[Project Page]{\url{https://visionxlab.github.io/IdeasHaveGenomes/}}
\hypersetup{
  pdftitle={Ideas Have Genomes: Benchmarking Scientific Lineage Reasoning and Lineage-Grounded Idea Generation},
  pdfauthor={Yifan Zhou, Qihao Yang, Yan Li, Donggang Li, Xiru Hu, Hokin Deng, Ziyang Gong, Xuanyi Zhou, Huacan Wang, Xiangchao Yan, Wanghan Xu, Wenlong Zhang, Shaofeng Zhang, Yue Zhou, Yifan Yang, Zhihang Zhong, Xue Yang},
  pdfsubject={Scientific lineage reasoning and lineage-grounded idea generation},
  pdfkeywords={scientific lineage, idea generation, benchmark, LLM evaluation}
}

\begin{document}
\maketitle
\vspace{-6mm}
\enlargethispage{1.5cm}
\begin{center}
\begin{tcolorbox}[
  colback=blue!3, colframe=blue!25,
  arc=6pt, boxrule=0.8pt,
  left=14pt, right=14pt, top=6pt, bottom=6pt,
  width=\textwidth
]
\itshape ``Descent with modification.'' \\[1pt]
\raggedleft \upshape --- Charles Darwin, \textit{On the Origin of Species} (1859)
\end{tcolorbox}
\end{center}
\vspace{-2pt}

\section{Introduction}
\label{sec:intro}

\begin{figure}[t]
\centering
\includegraphics[width=\textwidth]{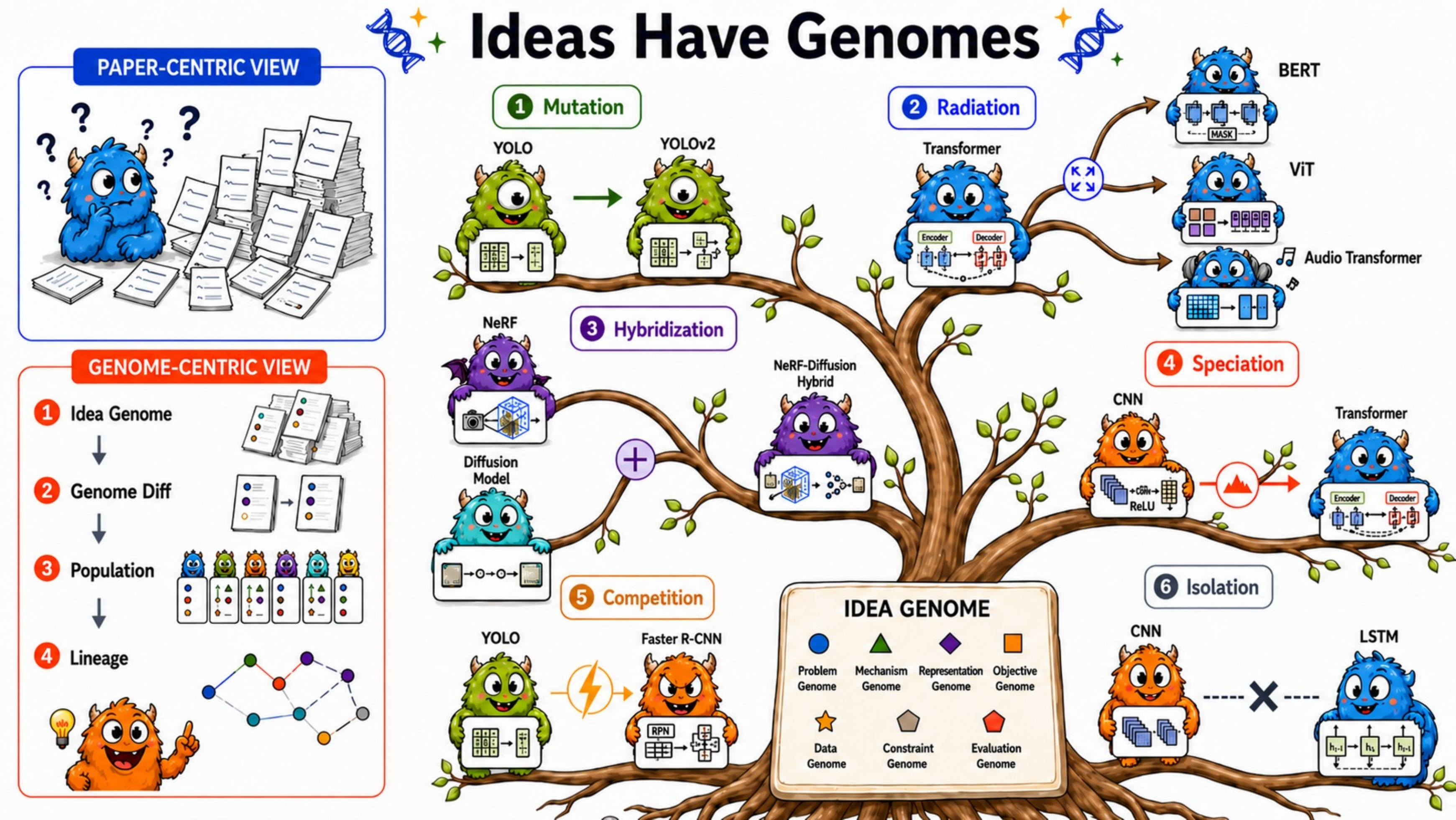}
\caption{\textbf{From paper-centric search to genome-centric lineage.} A paper-centric view leaves a model with many related papers but no explicit unit for deciding which ideas are inherited, mutated, recombined, or merely co-located. The genome-centric view first extracts \genome{} objects, then aligns them through \genomediff{} records into populations and lineages, making evolutionary relations explicit enough to evaluate.}
\label{fig:teaser}
\end{figure}
\enlargethispage{-1.2\baselineskip}

LLM-based auto-research systems now search literature, synthesize hypotheses, run experiments, and write paper-like reports~\citep{aiscientist,aiscientistv2,airesearcher}. Existing evaluations emphasize retrieval quality, factuality, writing fluency, novelty, workflow competence, or pairwise preference~\citep{litsearch,scieval,sgibench,scievalkit,scimon,si2024,zheng2024judging}. These criteria leave a harder question mostly untested: when a proposal claims to extend a research direction, does it inherit the right mechanism, repair the right limitation, and stay coherent with the lineage it builds on? Figure~\ref{fig:teaser} motivates this shift from paper-centric context gathering to genome-centric lineage comparison.

The distinction matters because scientific progress is not the same as topical proximity. Two papers can share a task without inheriting from one another, and two papers can look textually distant while carrying the same core mechanism forward. YOLOv2~\citep{redmon2017yolov2}, for example, modifies YOLO's~\citep{redmon2016yolo} single-shot detection mechanism through anchor boxes, batch normalization, and multi-scale training: the lineage is visible as local mutation. DETR~\citep{carion2020detr} also studies object detection, but it replaces the Faster R-CNN~\citep{ren2015faster} pipeline with set prediction and a Transformer architecture. It shares the task ecology, not the anchor-based driver mechanism. Titles, abstracts, citation edges, and embeddings tend to blur these cases together. A model can retrieve the right papers and still miss the parent mechanism or repaired limitation; a generated idea can sound novel while inheriting no coherent mechanism.

Evaluating this behavior requires a representation below the paper level, one that makes idea inheritance explicit enough to audit. We call the target capability \emph{scientific lineage competence}: the ability to (i)~abstract a paper into \genome{} objects, (ii)~trace which \genome{} objects persist, mutate, or disappear across papers, (iii)~explain the dominant dynamics of a transition, (iv)~verify whether a proposed lineage is coherent rather than merely topical, and (v)~generate a proposal that can be inserted into an existing lineage as a plausible descendant.

\ideagene{} turns this capability into an evaluable object. It represents each paper or proposal as a set of \textbf{\genome{}} objects, where each \genome{} is a minimal, typed, evidence-grounded idea structure used for lineage comparison. A \textbf{\genomediff{}} aligns \genome{} objects across predecessor and successor work, recording inheritance, mutation, loss, external import, and novel insertion. This is an operational choice rather than a biological theory of science: the framework fixes the granularity and evidence contract needed for comparison.

We instantiate the framework as \textbf{\igbenchfull{} (\igbench{})}. The benchmark contains 1,961 golden lineage traces across 10 scientific domains, 1,085 curated \genome{} objects, and 920 pairwise \genomediff{} records. It has two parts. \geneexam{} is a closed-form benchmark with 42 task types and 1,029 instances, testing \genome{} abstraction, inheritance tracing, evolutionary reasoning, and lineage verification. \genearena{} is an open-ended generation evaluation: systems write proposals under three controlled information settings---Question-only, Library, and Lineage---and are scored by a lineage-conditioned Population-Evolution Score (PES). PES asks whether a proposal can be inserted as a coherent descendant in a specified lineage population: inheriting the right \genome{} objects (Heredity), introducing meaningful variation relative to neighboring work (Variation), and offering selection value for future research (Selection).

Experiments on 14 LLM-based scientists---direct LLMs, research-agent frameworks, and CLI harnesses---show that plausible research text is not the same as lineage competence. On \geneexam{}, the best system reaches only 27.3\% exact accuracy. The common failure is compositional: models often recover local signals but fail to keep parent choice, driver assignment, object fate, and verification flags jointly consistent. In \genearena{}, structured lineage context does not simply raise every score; it reshuffles the ranking, separating systems that can use lineage evidence from those that only benefit from more text.

Our contributions are:
\begin{enumerate}[noitemsep,topsep=2pt]
\item \textbf{\ideagene{} framework for scientific lineage.} We define \genome{}, genome extraction, and \genomediff{}, together with six operational evolutionary dynamics (mutation, adaptive radiation, hybridization, speciation, niche competition, isolation) for aligning, classifying, and auditing idea inheritance across papers.
\item \textbf{\igbench{} benchmark.} We build a two-part benchmark: \geneexam{} for closed-form lineage reasoning (42 task types, 1,029 instances) and \genearena{} for lineage-grounded idea generation with a PES metric under controlled information settings.
\item \textbf{Empirical findings.} We show that frontier systems face a compositional bottleneck in lineage reasoning, that structured lineage evidence separates systems rather than uniformly helping them, and that generated ideas often sound plausible before they are lineage-coherent.
\end{enumerate}

\section{Related Work}
\label{sec:related}

\gene{} sits between several evaluation traditions. Scientific lineage needs both a representation layer and a benchmark: paper-level relevance alone is too coarse, while generation-only preference is too unconstrained.

\textbf{Scientific retrieval and QA.}
Scientific retrieval and representation benchmarks such as LitSearch and SciRepEval~\citep{litsearch,scirepeval} test whether systems can find or embed relevant papers; sentence and document embeddings~\citep{reimers2019sentence,wang2022e5} and multi-document scientific summarization~\citep{lu2020survey} add complementary representation tasks. Scientific QA, reasoning, general-intelligence, and end-to-end autonomous-research suites~\citep{scieval,sgibench,scievalkit,researchclawbench,sciq,arb} test factual, workflow, and problem-solving competence. \gene{} asks a different question: once the relevant work is available, can a system identify which idea structures are actually transmitted across papers?

\textbf{Automated research and ideation.}
Recent systems automate literature search, hypothesis generation, experimentation, and paper drafting~\citep{aiscientist,aiscientistv2,airesearcher,agentlab,coiagent}. Other work studies novelty-aware generation~\citep{scimon,nova,deepideation}, iterative ideation from literature~\citep{researchagent,si2024}, retrieval-augmented synthesis~\citep{openscholar,storm}, and multi-agent or fully autonomous scientific collaboration~\citep{sciagents,autoresearchclaw}. These systems make lineage-grounded evaluation more urgent. A proposal can be fluent and literature-aware, yet still fail to inherit the parent mechanism or repair the stated limitation of the work it claims to extend.

\textbf{Science of science and citation structure.}
Science-of-science work maps fields through citations, communities, and paradigm shifts~\citep{fortunato2018science,jurgens2018,kuhn1962}. Citation-intent classification~\citep{cohan2019} moves closer to functional roles, but still works at the paper or sentence level rather than at \genome{} alignment. Recent method-evolution infrastructure such as Intern-Atlas~\citep{internatlas} also moves beyond document-level citation topology by representing method-level entities and lineage relations. The evolutionary metaphor has deep roots~\citep{darwin1859}, yet computational uses often remain informal. \gene{} treats citation and time as candidate evidence, while making the evaluative object the \genomediff{} alignment itself.


\textbf{LLM evaluation methodology.}
LLM-as-judge~\citep{zheng2024judging} and Chatbot Arena~\citep{chiang2024chatbotarena} make model-judge and ELO protocols scalable. Work on position preference~\citep{wang2023fair} and length preference~\citep{dubois2024length} shows why judge design must be controlled, and holistic frameworks~\citep{liang2023helm,lin2024wildbench} motivate multi-dimensional scoring. Agent benchmarks such as SWE-bench~\citep{swebench}, GAIA~\citep{gaia}, and $\tau$-bench~\citep{tbench} evaluate tool use, but not scientific lineage competence. \genearena{} builds on this methodology while conditioning judgments on genome-centric lineage packets and neighboring populations.

\begin{table}[t]
\centering
\small
\caption{Positioning of \gene{} relative to adjacent evaluation paradigms.}
\label{tab:related_positioning}
\begin{tabular*}{\textwidth}{@{\extracolsep{\fill}}>{\raggedright\arraybackslash}p{3.8cm}>{\raggedright\arraybackslash}p{2.4cm}c c c c@{}}
\toprule
\textbf{Paradigm} & \textbf{Unit} & \textbf{Genome?} & \textbf{Diff?} & \textbf{Understand?} & \textbf{Generate?} \\
\midrule
Paper retrieval / QA & Paper or passage & \nomark & \nomark & Factual & \nomark \\
Scientific embeddings & Paper vector & \nomark & \nomark & Similarity & \nomark \\
Citation graph analysis & Citation edge & \nomark & Partial & Structural & \nomark \\
Literature-grounded ideation & Paper summary & Partial & Implicit & Weak & \yesmark \\
Automated research agents & Workflow trace & No fixed unit & \nomark & Indirect & \yesmark \\
\textbf{\gene{}} & \genome{} & \yesmark & \yesmark & \yesmark & \yesmark \\
\bottomrule
\end{tabular*}
\begin{flushleft}
\footnotesize Automated research agents include AI Scientist-v2, AI-Researcher, Agent Laboratory, CoI-Agent, SciAgents, and AutoResearchClaw.
\end{flushleft}
\end{table}

\section{IdeaGene Framework}
\label{sec:framework}

The evolutionary analogy gives lineage evaluation a useful vocabulary~\citep{darwin1859}, but \igbench{} does not require a broad biological ontology of science. It uses a smaller operational layer: \genome{} objects, genome extraction, and \genomediff{} alignments. These three pieces make inheritance explicit enough to evaluate.

\subsection{Scientific Lineage Competence}
\label{sec:lineage_competence}

\emph{Scientific lineage competence} is the ability to identify, verify, and extend the mechanism-level inheritance structure behind scientific work. A lineage-competent system should abstract a paper into heritable \genome{} objects, trace which objects persist or change across papers, explain the dominant dynamics of a transition, reject merely topical links, and generate a proposal that can enter an existing lineage as a plausible descendant.

\subsection{Idea Genome}
\label{sec:formalism}

Within \ideagene{}, an \textbf{\genome{}} is the minimal auditable structure used for lineage evaluation: a typed, evidence-grounded, lineage-relevant idea object extracted from a paper or proposal. We represent a paper or proposal as a set of \genome{} objects:
\begin{equation}
G(p)=\{g_i=(t_i,z_i,e_i,c_i)\}_{i=1}^{m_p}.
\label{eq:idea-genome}
\end{equation}
Each \genome{} object $g_i$ has a role type $t_i \in \{\texttt{niche}, \texttt{mechanism}, \texttt{observation}, \texttt{limitation}, \texttt{delta}, \texttt{claim}\}$, a content description $z_i$, an evidence pointer $e_i$, and optional constraints $c_i$. The role type states what the object does in lineage reasoning: \texttt{niche} marks the problem environment, \texttt{mechanism} marks an inheritable method or design, \texttt{observation} marks a motivating empirical pattern, \texttt{limitation} marks a defect or bottleneck, \texttt{delta} marks the repair or design change relative to prior work, and \texttt{claim} marks the asserted outcome.

\subsection{Genome Extraction}
\label{sec:genome_extraction}

Genome extraction is the abstraction operator $\mathcal{E}: p \mapsto G(p)$ that converts a paper or proposal into auditable \genome{} objects. The operator is narrower than general summarization: it keeps the objects that will later support \genomediff{} alignment and lineage verification. A valid \genome{} satisfies four constraints:

\begin{itemize}[noitemsep,topsep=2pt,leftmargin=1.2em]
\item \textbf{Typed.} Each \genome{} has a functional role such as \texttt{mechanism}, \texttt{limitation}, or \texttt{delta}. This keeps the representation from becoming an unstructured summary and lets later alignments distinguish inherited mechanisms from repaired limitations.
\item \textbf{Evidence-grounded.} Each \genome{} points to textual or structural evidence in the source paper or proposal, such as a section, paragraph, sentence span, figure, table, or equation. Light abstraction is allowed; unsupported background knowledge or annotator inference is not.
\item \textbf{Minimally self-contained.} Each \genome{} is small enough to be inherited, mutated, lost, or recombined independently, but complete enough to express one functional idea. If two components can disappear or transfer independently in later work, they should be split; if one is unintelligible without the other, they should remain together.
\item \textbf{Lineage-relevant.} Each \genome{} must matter for a lineage judgment. We extract a detail only if inheriting, losing, changing, or importing it would change whether a successor is a coherent descendant. Otherwise it remains metadata.
\end{itemize}

\subsection{\genomediff{}}
\label{sec:genomediff}

Given a predecessor $p_s$ and successor $p_t$, a \textbf{\genomediff{}} $\Delta_{s\to t}$ aligns \genome{} objects in $G(p_s)$ to \genome{} objects in $G(p_t)$ by type and semantic role. Source objects are marked \textsc{Inherited}, \textsc{Mutated}, or \textsc{Lost}; unaligned target objects are \textsc{Novel} or \textsc{External}. Each \genomediff{} record also stores the primary transition driver, the relation to the surrounding task or domain setting, and an evidence-backed rationale. We construct these records through LLM-assisted extraction followed by expert audit.

\paragraph{Lineage versus co-location.}
The surrounding task, benchmark, dataset convention, and community form an \emph{EcologyContext}: they explain why papers occupy the same research environment, but they do not by themselves establish descent. Genome continuity is what makes a lineage claim. Shared setting without driver inheritance is treated as niche competition; inherited mechanisms moving into a new setting are treated as adaptive radiation.

\subsection{Evolutionary Dynamics}
\label{sec:dynamics}

Evolutionary dynamics classify \genomediff{} patterns into operational categories. The first check is driver inheritance. Without evidence for it, the relation is co-located rather than lineage.

\begin{table}[t]
\centering
\footnotesize
\caption{Evolutionary dynamics as operational GenomeDiff criteria. Each dynamics type corresponds to a specific alignment pattern between two \genome{} objects.}
\label{tab:dynamics}
\begin{tabular*}{\textwidth}{@{\extracolsep{\fill}}%
                >{\raggedright\arraybackslash}p{2.25cm}%
                >{\centering\arraybackslash}p{1.45cm}%
                >{\raggedright\arraybackslash}p{4.75cm}%
                >{\raggedright\arraybackslash}p{5.05cm}@{}}
\toprule
\textbf{Dynamics} & \textbf{Lineage?} & \textbf{GenomeDiff criterion} & \textbf{Canonical example} \\
\midrule
Mutation & Yes & Driver mechanism is inherited or locally mutated; niche remains same or nearby. & YOLO~\citep{redmon2016yolo} $\rightarrow$ YOLOv2~\citep{redmon2017yolov2}: one-stage detection persists; anchors, batch normalization, and multi-scale training repair local limits. \\
Adaptive Radiation & Yes & Driver mechanism persists, but moves into a new task, domain, or evaluation ecology. & Transformer~\citep{attention2017} $\rightarrow$ ViT~\citep{vit2020}: self-attention is inherited, but moved from token sequences to image patches. \\
Hybridization & Yes & Successor imports driver objects from two or more distinct lineages. & CLIP-style visual encoder~\citep{clip2021} + instruction-tuned LLM $\rightarrow$ LLaVA~\citep{llava}: visual alignment and chat behavior are imported from two lineages. \\
Speciation & Yes & Same or nearby niche, but the predecessor's driver mechanism is replaced by a new lineage-forming mechanism. & Faster R-CNN~\citep{ren2015faster} $\rightarrow$ DETR~\citep{carion2020detr}: CNN region proposals are replaced by Transformer set prediction for detection. \\
Niche Competition & No & Same ecology or problem niche, but no driver inheritance. & Faster R-CNN~\citep{ren2015faster} vs. YOLO~\citep{redmon2016yolo}: both solve object detection, but region proposals and one-stage regression are competing mechanisms. \\
Isolation & No & Neither shared ecology nor driver inheritance. & BERT~\citep{devlin2019bert} vs. YOLO~\citep{redmon2016yolo}: language understanding and object detection share neither task ecology nor driver mechanism. \\
\bottomrule
\end{tabular*}
\end{table}

Ambiguous cases use a fixed priority rule: Hybridization before Speciation, Speciation before Niche Competition when lineage evidence exists, and Adaptive Radiation before Mutation when the setting shift is the driver. These evolutionary dynamics are operational categories for consistent evaluation; they do not claim to exhaust every possible pattern of scientific development.

\section{\gene{}: Dataset and Evaluation Artifact}
\label{sec:benchmark}

\begin{figure}[t]
\centering
\includegraphics[width=\textwidth]{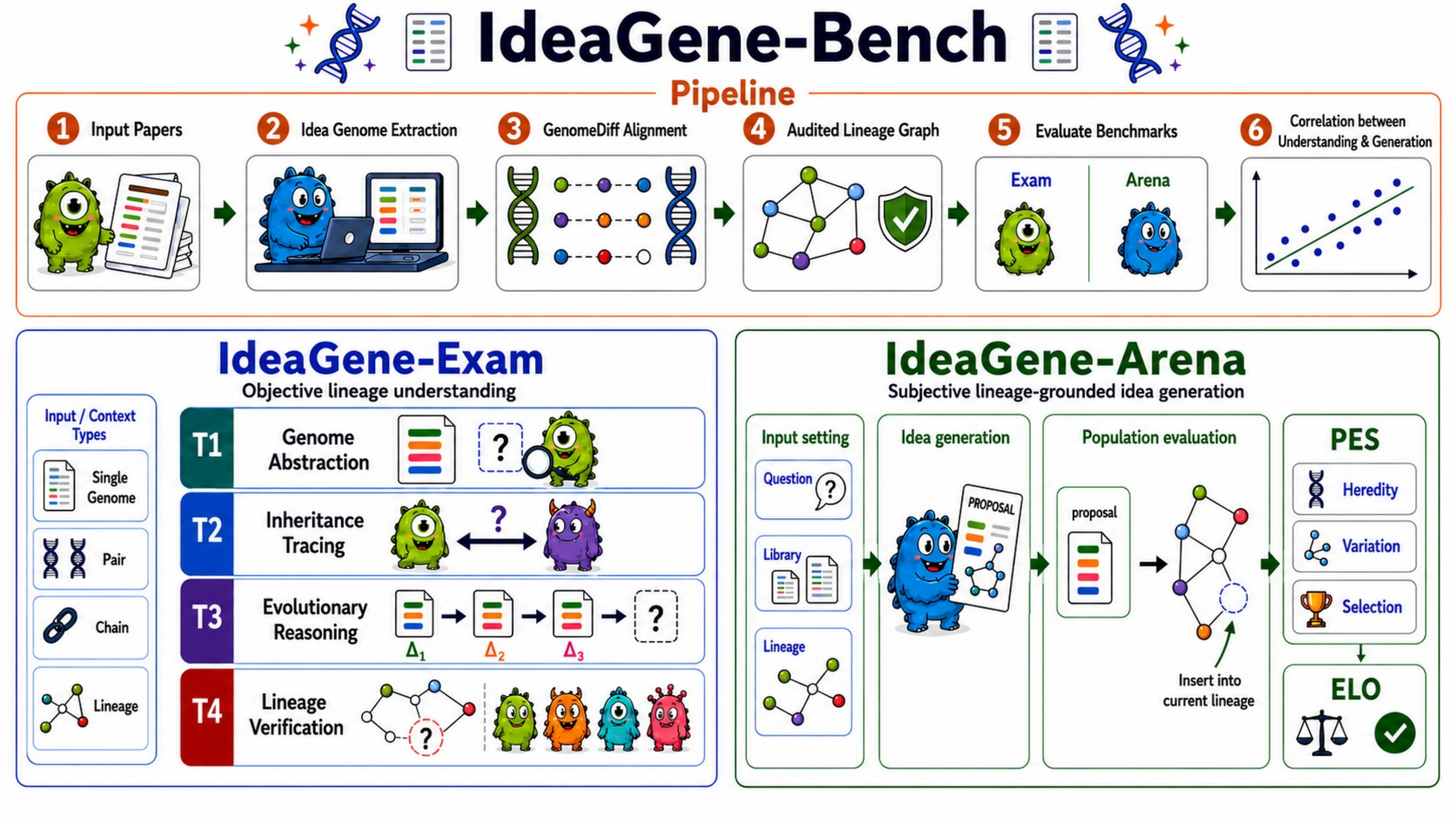}
\caption{\textbf{\gene{} evaluation design.} \gene{} converts input papers into a lineage substrate of audited \genome{} objects and \genomediff{} records, then evaluates two complementary capabilities: \geneexam{} for closed-form lineage understanding and \genearena{} for lineage-grounded idea generation. The same substrate supports the correlation between understanding and generation.}
\label{fig:eval}
\end{figure}

\gene{} instantiates the \ideagene{} framework as a reusable benchmark. It has two parts: \geneexam{} for closed-form lineage understanding and \genearena{} for lineage-grounded idea generation (Figure~\ref{fig:eval}).

\subsection{Data Construction and Quality Assurance}
\label{sec:data_construction}

\gene{} contains 1,961 golden lineage traces across 10 scientific domains---including NLP~\citep{devlin2019bert,radford2019language}, computer vision~\citep{redmon2016yolo,carion2020detr,vit2020}, multimodal learning~\citep{clip2021,llava}, and six additional domains (biology, chemistry, physics, materials, medicine, mathematics). These traces cover 1,085 curated \genome{} objects and 920 pairwise \genomediff{} records. Construction has four stages:

\begin{enumerate}[noitemsep,topsep=2pt,leftmargin=1.2em]
\item \textbf{Seed collection.} Experts nominate landmark and frontier papers in each domain.
\item \textbf{Trace expansion.} We expand candidate predecessors and successors through citation links, semantic search, and domain curation, producing lineage traces of 3--7 papers.
\item \textbf{Genome extraction and diff alignment.} Multi-pass LLM-assisted extraction converts each paper into typed \genome{} objects, which experts then audit. Pairwise \genomediff{} records add object alignment, fate annotation, and driver labels.
\item \textbf{Benchmark-level audit.} Programmatic checks and held-out annotators verify schema validity, answer contracts, temporal consistency, anonymization leakage, and trace coherence.
\end{enumerate}

\paragraph{Quality assurance.} We recruited 50 graduate annotators (master's and doctoral students across computer science, biology, physics, materials science, and other disciplines). They validated three points: \genomediff{} relation labels and role-type assignments during construction, \geneexam{} item difficulty through stratified human solving, and \genearena{} pairwise battles. Construction disagreements were adjudicated by a third annotator. Low human accuracy on \geneexam{} was checked to reflect compositional difficulty rather than label noise, and human judges reached 80\% agreement with the model-judge panel in \genearena{}. Inter-annotator agreement on dynamics labels is 84.7\% before adjudication.

\subsection{\geneexam{}: Lineage Understanding}
\label{sec:gene_exam}

\geneexam{} contains 42 main-challenge task types and 1,029 instances. Each item is
\begin{equation}
\tau=(c,x,\mathcal{A},y,m),
\label{eq:geneexam-item}
\end{equation}
where $c$ is the capability axis, $x$ is anonymized context, $\mathcal{A}$ is the answer space, $y$ is the gold target, and $m$ stores metadata. Scoring uses exact match: every required field must be correct at the same time. A lineage answer is not reliable if it identifies the right parent but assigns the wrong driver or object fate.

\begin{table}[t]
\centering
\small
\caption{\geneexam{} main challenge: 42 task types, 1,029 instances, four capability axes.}
\label{tab:benchmark}
\begin{tabular}{@{}>{\raggedright\arraybackslash}p{4.85cm}@{\;\;}c@{\;\;}c@{\;\;}>{\raggedright\arraybackslash}p{6.35cm}@{}}
\toprule
\textbf{Capability} & \textbf{Tasks} & \textbf{Inst.} & \textbf{Representative tasks} \\
\midrule
T1 Genome Abstraction & 5 & 125 & Identify field type, driver \genome{}, contribution role, or lineage position \\
T2 Inheritance Tracing & 12 & 313 & Reconstruct ordered lineages, align inherited \genome{} objects, match limitations to deltas \\
T3 Evolutionary Reasoning & 17 & 409 & Infer dynamics, driver, object fates, hybrid provenance, or multi-hop changes \\
T4 Lineage Verification & 8 & 182 & Detect intruders, wrong steps, missing links, citation conflicts, or parent mismatch \\
\bottomrule
\end{tabular}
\end{table}

T1 tests whether a model can read a single \genome{}. T2 asks it to trace inheritance across multiple \genome{} objects. T3 requires an explanation of a transition under \genomediff{} criteria. T4 verifies proposed lineage claims and is the closest closed-form analogue to idea generation, because a generated descendant must pass the same parent, coherence, and evidence checks.

\subsection{\genearena{}: Lineage-Grounded Generation}
\label{sec:gene_arena}

\genearena{} evaluates open-ended proposals under controlled information settings. Each instance contains a frontier question, a research domain, an optional paper library, an optional structured lineage, and a scoring rubric. The three settings separate different sources of capability:
\begin{itemize}[noitemsep,topsep=2pt,leftmargin=1.2em]
\item \textbf{Question}: domain and frontier question only; tests parametric ideation.
\item \textbf{Library}: unordered paper summaries; tests paper-centric context use.
\item \textbf{Lineage}: ordered trace with \genome{} objects and \genomediff{} evidence; tests lineage-structured context use.
\end{itemize}

\paragraph{Population-Evolution Score (PES).}
The primary metric is PES, a lineage-conditioned population-insertion score. Rather than judging an idea in isolation, PES evaluates each generated proposal $x$ as a candidate descendant of an ordered lineage $\mathcal{L}$ and a neighboring population $\mathcal{P}$ of related papers or proposals. The judge panel receives a lineage-structured context packet: the frontier question, the ordered lineage, relevant \genome{} objects, \genomediff{} evidence, neighboring population items, and the candidate proposal. PES then decomposes insertion quality into three dimensions that mirror Darwin's conditions for evolution by natural selection~\citep{darwin1859}:

\begin{itemize}[noitemsep,topsep=2pt,leftmargin=1.2em]
\item \textbf{Heredity} ($H$): Relative to $\mathcal{L}$, does the proposal inherit and build on the right parent \genome{} objects? This measures source continuity, mechanism preservation, and limitation-delta coherence under the \genomediff{} evidence.
\item \textbf{Variation} ($V$): Relative to $\mathcal{P}$, does the proposal introduce meaningful novelty? Cosmetic recombination is penalized; substantive mutation, transfer, or new \genome{} objects are rewarded.
\item \textbf{Selection} ($S$): Within the stated lineage population, is the proposal competitively viable? This measures feasibility, fit to the research environment, and whether the proposed work opens productive downstream directions.
\end{itemize}

\noindent Each dimension is scored on a 0--100 scale by a judge panel (3 model judges with position randomization) conditioned on $(\mathcal{L},\mathcal{P})$, and PES is the arithmetic mean:
\begin{equation}
\text{PES}(x \mid \mathcal{L},\mathcal{P}) =
\frac{1}{3}\left(H(x \mid \mathcal{L}) + V(x \mid \mathcal{P}) + S(x \mid \mathcal{L},\mathcal{P})\right).
\label{eq:pes}
\end{equation}
We report PES as the primary metric because it measures lineage-conditioned insertion quality rather than standalone appeal. Pairwise ELO~\citep{elo1978rating}, computed from judge-panel battles under a Bradley--Terry model~\citep{bradley1952rank}, is a complementary preference diagnostic. ELO and PES can diverge: fluency-oriented preferences may favor a polished but lineage-incoherent proposal over a less polished but better-grounded one. We validate PES reliability through inter-judge agreement (Krippendorff's $\alpha = 0.74$~\citep{krippendorff2011alpha}) and human concordance (80\% agreement with the model-judge panel on pairwise rankings).


\section{Experiments}
\label{sec:results}

\subsection{Evaluation Setup}


\paragraph{Systems and settings.}
\geneexam{} uses the 42-task, 1,029-instance main-challenge profile after removing calibration diagnostics that reached ceiling performance in pilot runs. \genearena{} covers 30 frontier tasks across 10 domains, yielding 1,260 generated proposals from 14 LLM-based scientists under Question, Library, and Lineage settings. Participants include eight direct LLMs, two GPT-5.5-based research agents, and four CLI harnesses that wrap GPT-5.5 or Claude Opus 4.7 in Codex or Claude Code workflows. This split lets us separate backbone ability, retrieval-heavy agent workflows, and lightweight tool-using scaffolds.

\paragraph{Metrics.}
\geneexam{} reports exact accuracy, so all required fields must be correct simultaneously. \genearena{} reports PES (Eq.~\ref{eq:pes}) as the primary lineage-conditioned population-insertion metric and ELO as a preference diagnostic. The three information settings form a controlled ablation: Question tests parametric ideation, Library tests paper-level context, and Lineage tests ordered \genome{} objects and \genomediff{} records.

\subsection{\geneexam{} Results}

\begin{table}[h]
\centering
\scriptsize
\caption{\textbf{\gene{} main leaderboard.} \geneexam{} measures closed-form lineage understanding with exact accuracy (\%). \genearena{} reports lineage-conditioned Population-Evolution Score (PES) and pairwise ELO as complementary diagnostics.}
\label{tab:gene-bench-main}
\resizebox{\textwidth}{!}{%
\begin{tabular}{@{}lrrrrrrrr@{}}
\toprule
\textbf{LLM-based Scientist} &
\multicolumn{5}{c}{\textbf{\geneexam{}}} &
\multicolumn{2}{c}{\textbf{\genearena{}}} \\
\cmidrule(lr){2-6}\cmidrule(lr){7-8}
 & \makecell[c]{\textbf{T1}\\Abstraction} &
\makecell[c]{\textbf{T2}\\Tracing} &
\makecell[c]{\textbf{T3}\\Reasoning} &
\makecell[c]{\textbf{T4}\\Verification} &
\textbf{Total} &
\textbf{PES} & \textbf{ELO} \\
\midrule
\rowcolor{gray!15} \multicolumn{8}{@{}l}{\textit{Direct LLMs}} \\
GPT-5.5 & 27.5 & 25.7 & 23.3 & 16.0 & 23.1 & 86.5 & 1213.2 \\
Claude Opus 4.7 & 28.5 & 21.9 & 17.1 & 14.5 & 19.3 & 82.6 & 1007.2 \\
Qwen3.6-Max-Preview & 26.9 & 22.5 & 18.8 & \textbf{17.4} & 20.6 & 79.8 & 882.7 \\
Gemini-3.1-pro-preview & 32.4 & 24.6 & 17.8 & 10.7  & 20.1 & 82.1 & 705.7 \\
Kimi-K2-Thinking & 27.9 & 22.1 & 19.6 & 14.8 & 20.4 & 81.1 & 793.9 \\
DeepSeek-V4-Pro & 23.9 & 20.6 & 18.6 & 8.5 & 17.9 & 82.7 & 918.8 \\
GLM-5.1 & 28.5 & 19.7 & 17.4 & 8.5 & 17.7 & 80.7 & 668.0 \\
MiniMax-M2.7 & 22.9 & 9.1 & 10.9 & 11.6 & 11.9 & 70.5 & 540.8 \\
\midrule
\rowcolor{gray!15} \multicolumn{8}{@{}l}{\textit{Research agents}} \\
AI Scientist v2 (GPT 5.5) & 28.1 & 26.9 & 22.3 & 15.1 & 23.0 & 80.6 & 1253.3 \\
CoI-Agent (GPT 5.5) & 26.9 & 27.4 & 22.4 & 13.4 & 22.7 & 80.5 & 1249.9 \\
\midrule
\rowcolor{gray!15} \multicolumn{8}{@{}l}{\textit{CLI harnesses}} \\
Codex (GPT-5.5) & \textbf{31.8} & 30.3 & 23.6 & 13.7 & 24.6 & \textbf{86.7} & \textbf{1476.1} \\
Claude Code (GPT-5.5) & 31.5 & \textbf{37.9} & \textbf{25.3} & 12.7 & \textbf{27.3} & 86.1 & 1420.6 \\
Codex (Claude Opus 4.7) & 34.4 & 29.7 & 17.0 & 10.5 & 21.8 & 82.7 & 900.8 \\
Claude Code (Claude Opus 4.7) & 27.9 & 26.3 & 21.0 & 13.8 & 22.0 & 83.3 & 1159.7 \\
\bottomrule
\end{tabular}
}
\end{table}

\begin{figure}[h]
\centering
\includegraphics[width=0.72\textwidth]{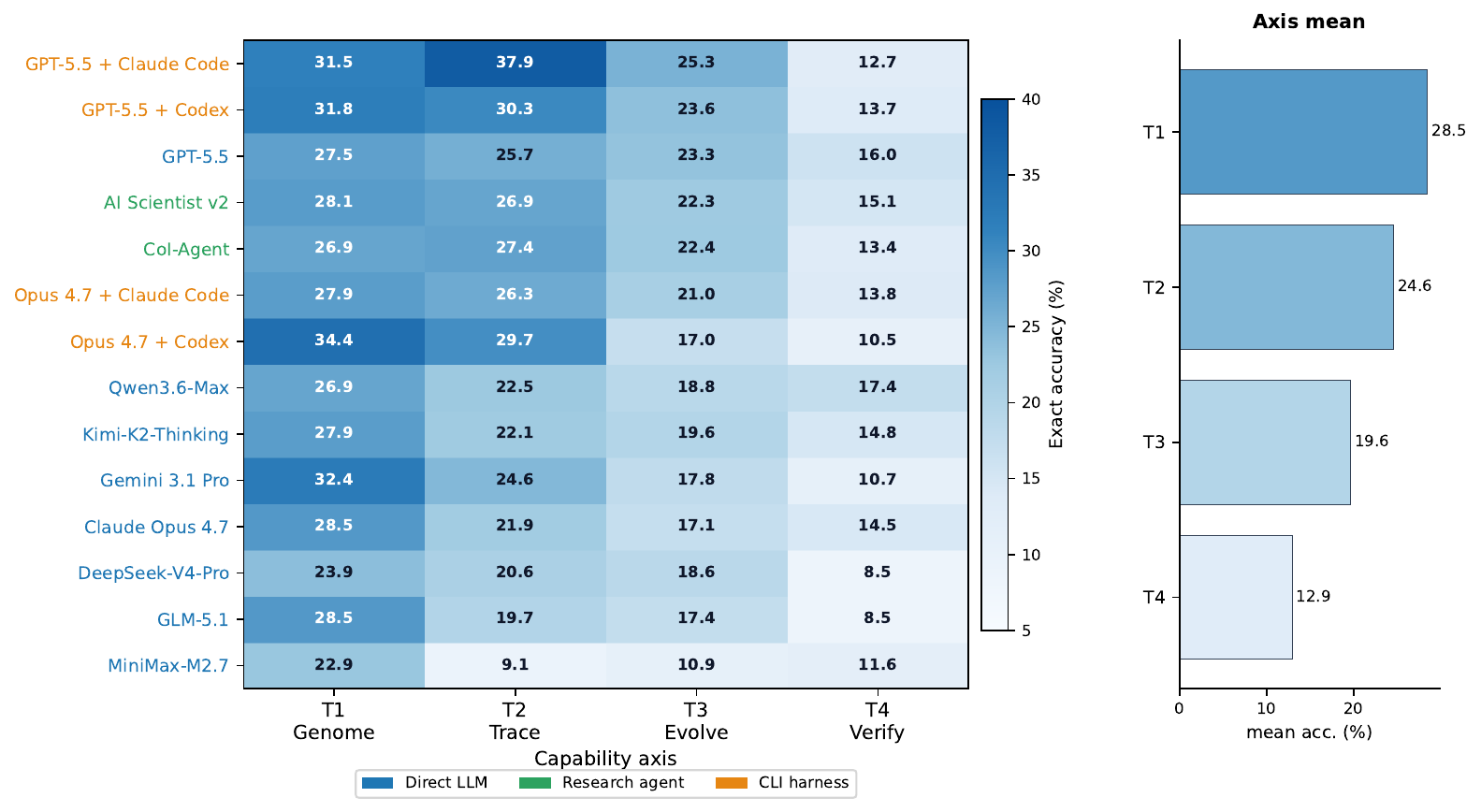}
\caption{\textbf{\geneexam{} exact-accuracy heatmap.} Exact accuracy across T1--T4 for all evaluated systems, with verification remaining the hardest axis.}
\label{fig:exam_bottleneck_app}
\end{figure}

\textbf{Lineage reasoning is compositionally hard.}
The strongest direct LLM, GPT-5.5, reaches 23.1\% exact accuracy; the best harness, GPT-5.5 + Claude Code, reaches 27.3\% (Table~\ref{tab:gene-bench-main}; Appendix~\ref{fig:exam_bottleneck_app}). Errors are rarely simple misses. A model may identify the parent paper but assign the wrong dynamics label, or infer the driver \genome{} but misclassify its fate. Exact scoring exposes the consistency failures that matter for downstream generation.

\textbf{Capability-axis breakdown.}
Performance falls from T1 Genome Abstraction (single-genome reading, best: 34.4\%) through T2 Inheritance Tracing (multi-genome tracing, best: 37.9\%) and T3 Evolutionary Reasoning (best: 25.3\%) to T4 Lineage Verification (best: 17.4\%). The T1$\to$T4 gradient tracks the added compositional burden. T1 requires reading one \genome{}; T4 requires parent identity, \genome{} compatibility, driver consistency, and evidence validity to hold together. T4 is also the capability closest to generation: a system that cannot verify lineage claims cannot reliably generate coherent descendants.

\textbf{Tool scaffolding helps retrieval more than consistency.}
CLI harnesses improve most on T2 Inheritance Tracing (GPT-5.5: 25.7\% $\rightarrow$ 37.9\% with Claude Code), where iterative tool use can retrieve and compare extracted \genome{} objects across papers. The gains shrink on T3 Evolutionary Reasoning and nearly vanish on T4 Lineage Verification. Current scaffolds therefore help information gathering more than compositional consistency checking. Research agents (AI Scientist v2 and CoI-Agent in the main ranking) stay close to direct GPT-5.5 on most axes, suggesting that retrieval-heavy workflows alone do not add lineage-reasoning capability.

\FloatBarrier

\subsection{\genearena{} Results}

\begin{figure}[t]
\centering
\begin{subfigure}[t]{0.56\textwidth}
\centering
\includegraphics[width=\linewidth]{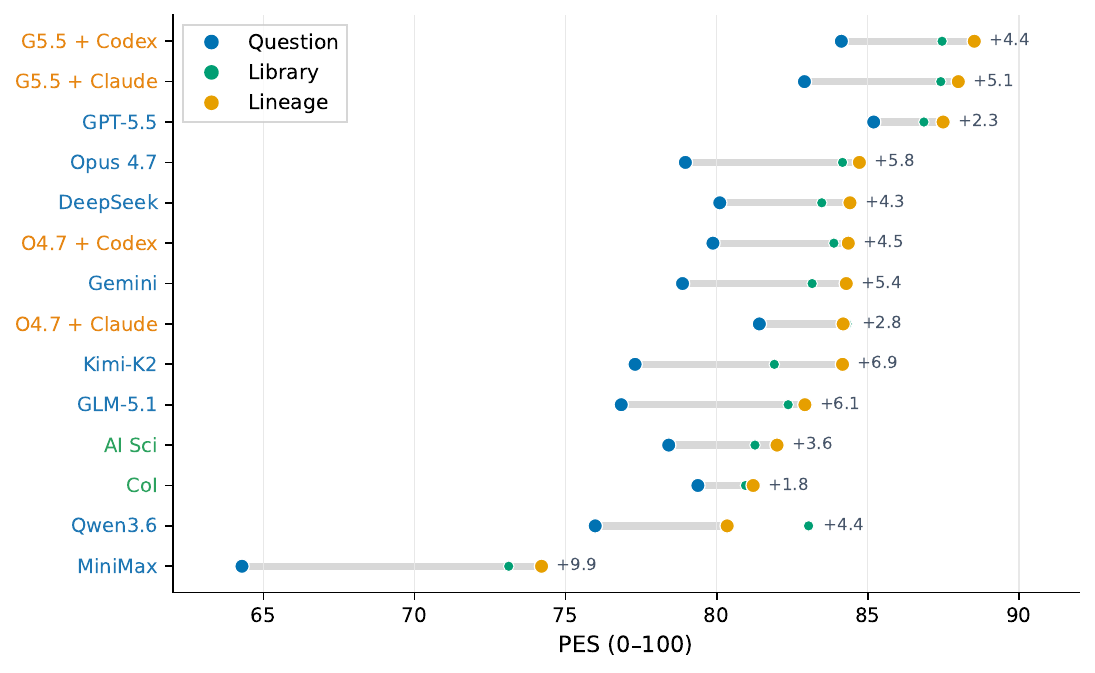}
\caption{PES across information settings.}
\end{subfigure}\hfill
\begin{subfigure}[t]{0.42\textwidth}
\centering
\includegraphics[width=\linewidth]{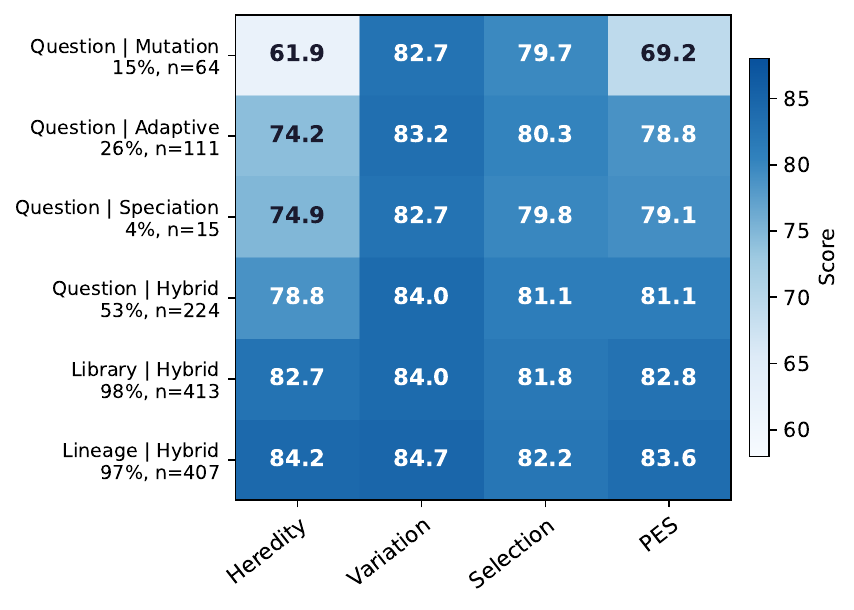}
\caption{PES decomposition by dynamics mode.}
\end{subfigure}
\caption{\textbf{\genearena{} PES analysis.} (a)~Question--Library--Lineage trajectories show that structured lineage context improves generation quality unevenly across systems. (b)~PES decomposed by setting--dynamics group, sorted by PES ascending. Only groups with $n{\ge}15$ are shown: Question produces four dynamics modes while Library and Lineage concentrate on Hybridization (${\ge}97$\%). Variation and Selection are nearly constant across all rows, while Heredity alone explains the PES gap---from 61.9 (Question|Mutation) to 84.2 (Lineage|Hybrid).}
\label{fig:pes_analysis}
\end{figure}

\textbf{Lineage context separates systems.}
The gain from Question to Lineage varies by system (Figure~\ref{fig:pes_analysis}a). GPT-5.5 already performs well under Question-only prompts (PES 85.2) and gains only +2.3 from lineage context. Weaker systems such as Kimi-K2-Thinking gain more (+6.9). The median gain is +4.4, so lineage context helps on average, but its main effect is diagnostic: it separates systems that can use \genome{} structure from those that cannot.

\begin{figure}[h]
\centering
\begin{subfigure}[t]{0.48\textwidth}
\centering
\includegraphics[width=\linewidth]{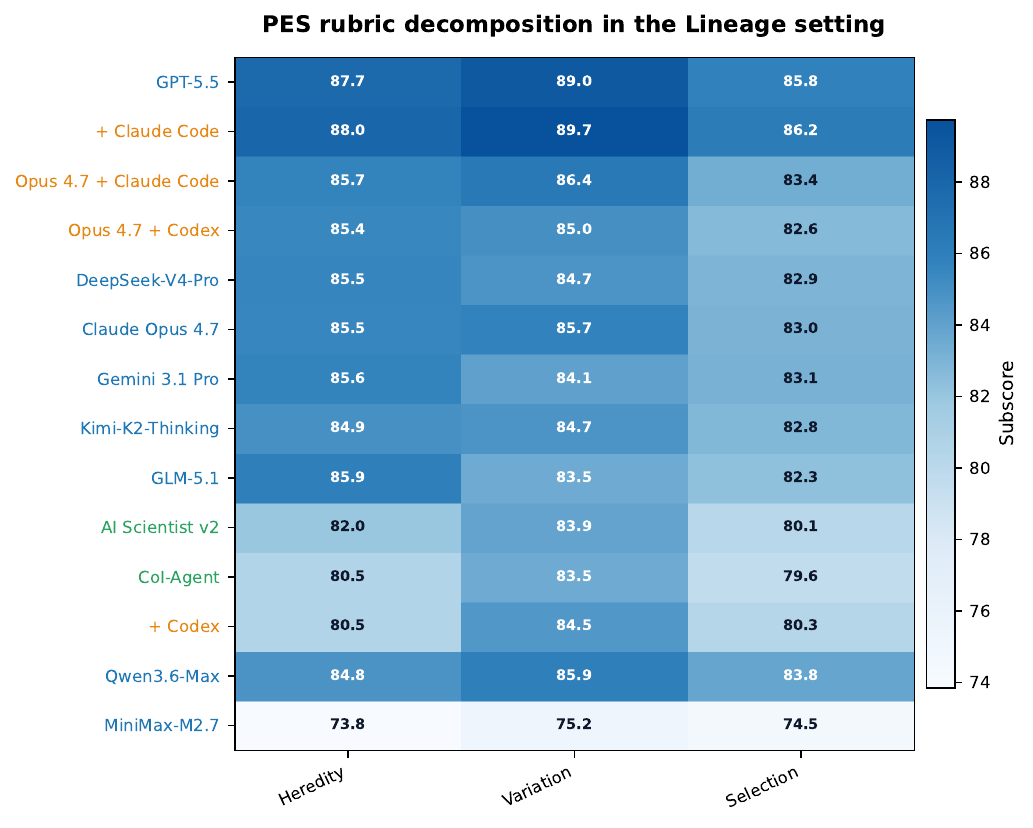}
\caption{PES rubric decomposition (H / V / S).}
\end{subfigure}\hfill
\begin{subfigure}[t]{0.48\textwidth}
\centering
\includegraphics[width=\linewidth]{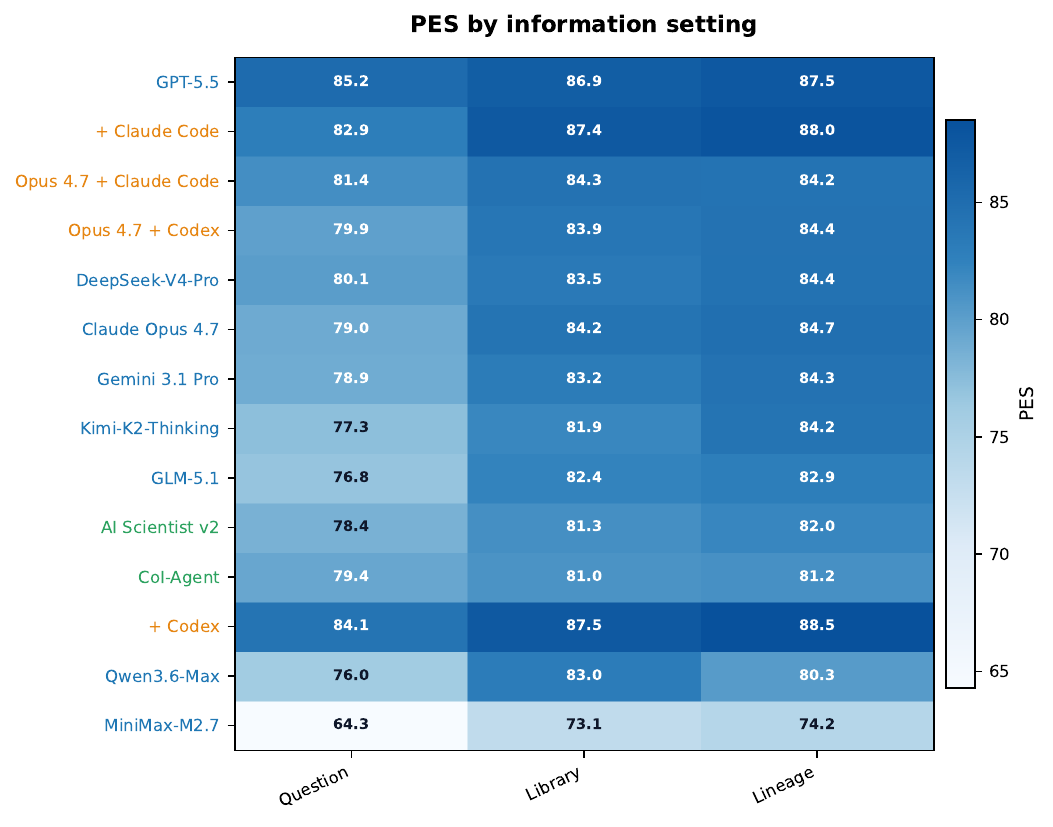}
\caption{PES by information setting (Question / Library / Lineage).}
\end{subfigure}
\caption{\textbf{PES decomposition and information-setting breakdown.} (a)~Heredity, Variation, and Selection sub-scores under the genome-centric scoring packet. Variation consistently exceeds Heredity across systems, confirming the plausibility--coherence gap identified in Finding~3. (b)~PES under Question, Library, and Lineage settings. Question$\to$Lineage gains are heterogeneous: GPT-5.5 moves only from 85.2 to 87.5, while weaker systems gain more from explicit lineage context.}
\label{fig:pes_dim_app}
\end{figure}

\textbf{The PES gain is Heredity-driven.}
Figure~\ref{fig:pes_analysis}b decomposes PES by setting--dynamics group. Variation and Selection stay nearly constant across rows (82.7--84.7 and 79.7--82.2), while Heredity explains the gap. Question-only Mutation---parametric invention without a valid parent mechanism---scores Heredity 61.9 and PES 69.2. Lineage-setting Hybridization---recombination anchored to explicit parent \genome{} objects---reaches Heredity 84.2 and PES 83.6. The Q$\to$Lineage improvement is therefore not about generating more novelty; Variation is already high. It is about grounding the proposal in the correct parent mechanism (Heredity $+$5.4 from Question|Hybrid to Lineage|Hybrid). Per-system H/V/S breakdowns (Appendix~\ref{fig:pes_dim_app}) show the same Variation-over-Heredity gap across all 14 participants.

\begin{figure}[h]
\centering
\includegraphics[width=0.58\textwidth]{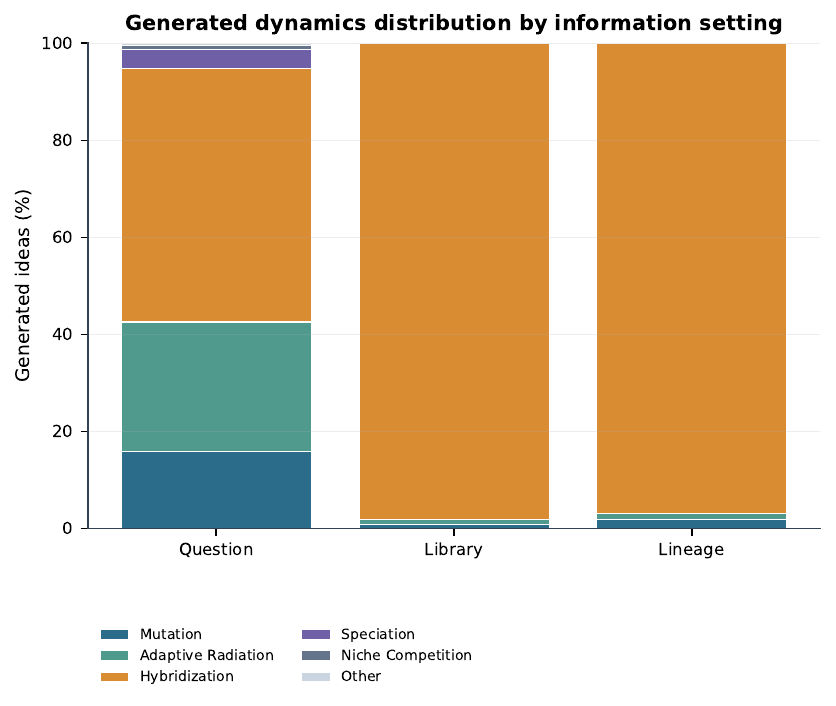}
\caption{\textbf{Generated dynamics distribution by information setting.} Stacked proportions of post-hoc dynamics labels for generated \genearena{} proposals. The main text replaces this pure frequency view with a quality map because the same dynamics label can have different Heredity and PES depending on whether lineage evidence is available.}
\label{fig:arena_dynamics_dist_app}
\end{figure}

\textbf{Dynamics labels only become meaningful when grounded in a lineage.}
Post-hoc dynamics classification shows that Question-only prompts produce a broad mixture of Mutation, Adaptive Radiation, and Hybridization, while Library and Lineage prompts concentrate around Hybridization (Appendix~\ref{fig:arena_dynamics_dist_app}). The label itself is not a quality score. Question-only Mutation has low PES because it invents a local modification without a valid parent mechanism. Lineage-setting Hybridization has high PES because the recombination is anchored to explicit parent \genome{} objects and \genomediff{} evidence. The dynamics label explains \emph{how} a proposal moves; PES asks whether that move is a coherent population insertion.

\textbf{PES versus ELO divergence.}
ELO rankings partially diverge from PES rankings (Spearman $\rho = 0.82$). The divergence appears most clearly when a fluent but lineage-incoherent proposal wins pairwise battles against a less polished but better-grounded one. This motivates PES as the primary metric: pairwise preference captures surface appeal, while PES measures lineage-grounded population insertion.

\subsection{Key Findings}

\begin{figure}[t]
\centering
\begin{subfigure}[t]{0.42\textwidth}
\centering
\includegraphics[width=\linewidth]{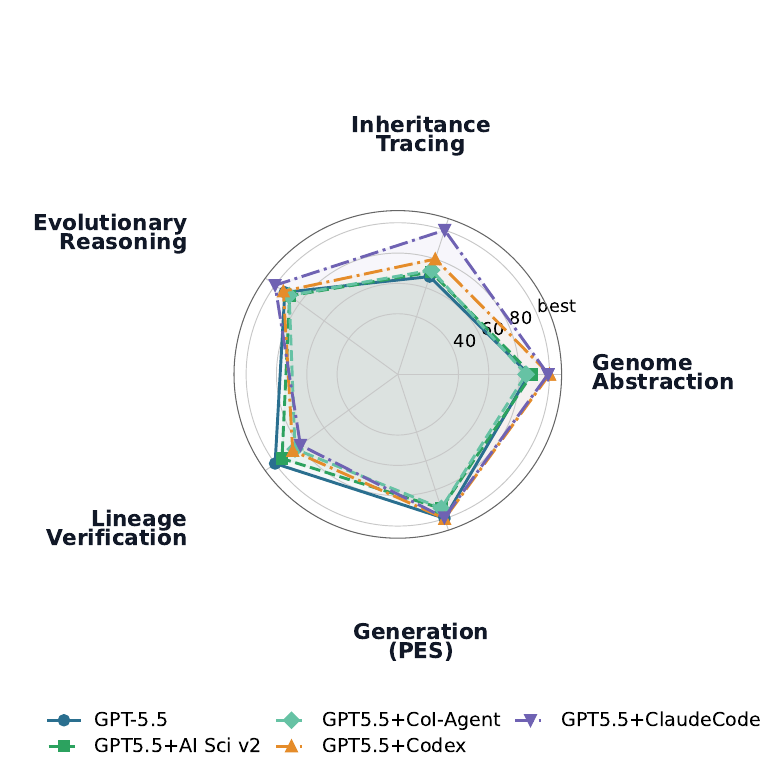}
\caption{Five capability dimensions.}
\end{subfigure}\hfill
\begin{subfigure}[t]{0.56\textwidth}
\centering
\includegraphics[width=\linewidth]{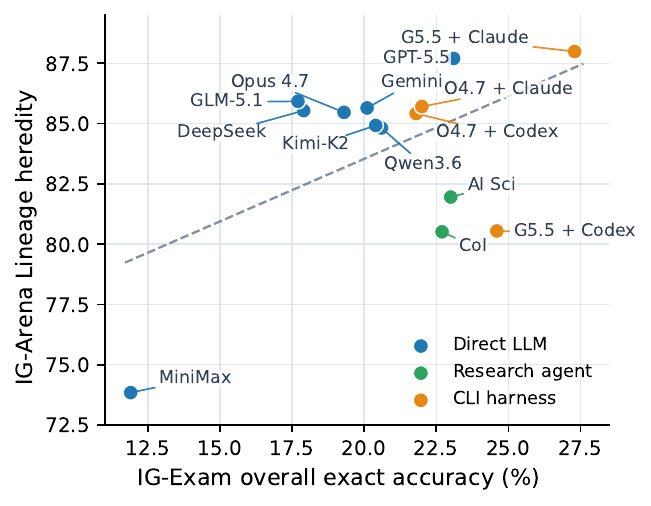}
\caption{Understanding--generation bridge.}
\end{subfigure}
\caption{\textbf{Capability profile and understanding--generation bridge.} (a)~Radar plot over five capability dimensions for five GPT-5.5-backbone systems---the direct LLM, two research agents (AI Scientist v2, CoI-Agent), and two CLI harnesses (Codex, Claude Code)---reducing backbone confounds. Agents nearly overlap with the direct LLM, while CLI harnesses redistribute strengths rather than uniformly improving every capability. (b)~Systems with stronger closed-form lineage understanding tend to preserve lineage-setting heredity better during generation; the moderate correlation shows the two evaluations remain complementary.}
\label{fig:capability_bridge}
\end{figure}

\textbf{Finding 1: Paper-level evidence is not enough.}
Library context provides more information than Question-only, but unordered paper summaries do not say which \genome{} objects are inherited, mutated, lost, or externally imported. The Lineage setting isolates that added structure. Its gain in lineage-conditioned PES, especially Heredity, shows that \genomediff{} structure carries signal beyond paper summaries.

\textbf{Finding 2: Verification bridges understanding and generation.}
The hardest \geneexam{} tasks are T4 Lineage Verification tasks, and \genearena{} needs the same checks: a proposal must identify a parent, inherit compatible \genome{} objects, repair a real limitation, and avoid invalid recombination. Figure~\ref{fig:capability_bridge}b plots each system's \geneexam{} exact accuracy against its Lineage-setting \genearena{} Heredity. The moderate positive association suggests that closed-form lineage understanding and open-ended generation quality are related, but not redundant.

\textbf{Finding 3: The backbone-controlled profile separates retrieval help from lineage competence.} Figure~\ref{fig:capability_bridge}a compares five GPT-5.5-backbone systems across five capability dimensions. Research agents (AI Scientist v2, CoI-Agent) nearly overlap with the direct LLM, so their retrieval-heavy workflows do not add much lineage-reasoning capability. CLI harnesses (Codex, Claude Code) substantially improve T2 Inheritance Tracing and maintain T1 Genome Abstraction, but show little gain on T3 Evolutionary Reasoning and even decrease T4 Lineage Verification accuracy. Agents also score lower on Generation (PES) than the direct LLM, suggesting that multi-step pipelines can hurt generation coherence. Tool scaffolds therefore amplify retrieval-dependent capabilities while leaving compositional reasoning largely unchanged---the bottleneck already visible in Table~\ref{tab:gene-bench-main}.

\textbf{Finding 4: Systems overproduce plausibility relative to lineage coherence.}
Generated ideas often sound useful before they preserve the exact parent mechanism or limitation-delta relation. The PES dynamics decomposition (Figure~\ref{fig:pes_analysis}b) shows why: Variation stays high across settings and dynamics modes, while Heredity drives the PES gap. Ungrounded combination can be high-variation but low-heredity; genome-grounded recombination can be novel and coherent at the same time.

\FloatBarrier

\section{Conclusion}
\label{sec:conclusion}

\gene{} reframes automated-research evaluation around scientific lineage. The central question is not only whether a system can write plausible research text, but whether it can identify, verify, and extend the inherited mechanisms that make a proposal a coherent descendant of prior work. The \ideagene{} framework---typed \genome{} objects, \genomediff{} records, and evolutionary dynamics---provides the representation layer for making that judgment precise and auditable.

The results make that distinction visible. Even the best system reaches only 27.3\% exact accuracy on \geneexam{}; tool scaffolds help T2 Inheritance Tracing but barely move T4 Lineage Verification; and structured lineage context separates systems rather than lifting them uniformly. In \genearena{}, PES decomposition shows the same plausibility--coherence gap: proposals can sound novel while failing to preserve the parent mechanism or limitation-delta relation needed for coherent descent. These findings point to a concrete design direction: auto-research systems need compositional verification modules, not only better retrieval. \gene{} provides an evaluation substrate for measuring whether future auto-research systems can move from retrieving papers and producing plausible text toward verifying and extending scientific lineages.

\section{Limitations}
The six evolutionary dynamics in \gene{} are operational categories for auditable evaluation, not an exhaustive theory of scientific development. Real lineages can mix multiple transition patterns within the same successor; \gene{} assigns primary drivers to keep annotation and evaluation consistent.


\bibliographystyle{plainnat}
\bibliography{references}


\clearpage
\beginappendix

\section{Main-Challenge Task Inventory}
\label{app:tasks}

\begin{table}[h]
\centering
\scriptsize
\caption{Main \geneexam{} challenge: 42 task types across 4 capability dimensions, totaling 1,029 instances. Calibration/local-matching diagnostics are retained in the code release but excluded from the paper leaderboard.}
\label{tab:all_tasks}
\begin{tabular}{@{}p{1.15cm}p{1.2cm}p{5.35cm}r@{}}
\toprule
\textbf{Axis} & \textbf{ID} & \textbf{Task Type} & \textbf{N} \\
\midrule
\multirow{5}{*}{T1} & T1-01 & Contribution Role & 30 \\
& T1-02 & Genome-Field Type & 25 \\
& T1-03 & Driver-Passenger Roles & 20 \\
& T1-04 & Lineage Position & 25 \\
& T1-05 & Cross-Lineage Bridge & 25 \\
\midrule
\multirow{12}{*}{T2} & T2-01 & Five-Genome Lineage Reconstruction & 25 \\
& T2-02 & Six-Genome Lineage Reconstruction & 25 \\
& T2-03 & Seven-Genome Lineage Reconstruction & 25 \\
& T2-04 & Lineage Grouping (8) & 25 \\
& T2-05 & Medium Lineage Grouping (8) & 25 \\
& T2-06 & Three-Lineage Grouping (9) & 25 \\
& T2-07 & Limitation-Delta Match & 25 \\
& T2-08 & Mixed Limitation-Delta Match & 25 \\
& T2-09 & Chained Limitation-Delta Match & 25 \\
& T2-10 & Two-Paper Genome-Field Assignment & 29 \\
& T2-11 & Three-Paper Genome-Field Assignment & 29 \\
& T2-12 & Genome Alignment & 30 \\
\midrule
\multirow{17}{*}{T3} & T3-01 & Single-Step Dynamics & 25 \\
& T3-02 & Genome-Field Fate & 25 \\
& T3-03 & Driver Dynamics & 25 \\
& T3-04 & Shown Genome-Field Fate & 22 \\
& T3-05 & Driver-Fate Summary & 23 \\
& T3-06 & Mechanism-Only Dynamics & 25 \\
& T3-07 & Blind Genome-Field Fate & 25 \\
& T3-08 & Unlabeled Driver Inference & 24 \\
& T3-09 & Relation Classification & 21 \\
& T3-10 & Directional Genome Choice & 25 \\
& T3-11 & Evolutionary Tempo & 30 \\
& T3-12 & Evolutionary Pattern & 25 \\
& T3-13 & Hidden Genome Fate & 24 \\
& T3-14 & Hybrid Provenance & 25 \\
& T3-15 & Multi-Hop Genome Tracking & 20 \\
& T3-16 & Dynamics Boundary & 25 \\
& T3-17 & Multi-Citation Relation & 20 \\
\midrule
\multirow{8}{*}{T4} & T4-01 & Genome Consistency Check & 25 \\
& T4-02 & Domain Intruder Detection & 25 \\
& T4-03 & Local Lineage Repair & 25 \\
& T4-04 & Next-Hop Prediction & 24 \\
& T4-05 & Parent-Genome Identification & 25 \\
& T4-06 & Missing-Link Recovery & 18 \\
& T4-07 & Genome-Bridge Validation & 20 \\
& T4-08 & Citation Consistency & 20 \\
\bottomrule
\end{tabular}
\end{table}

\begin{table}[h]
\centering
\scriptsize
\caption{Active \genearena{} task inventory for the main run: 30 tasks balanced across 10 scientific domains.}
\label{tab:traces}
\begin{tabular}{p{2.6cm}p{8.1cm}c}
\toprule
\textbf{Domain} & \textbf{Task Type} & \textbf{Count} \\
\midrule
Computer Science & AgentFramework, LLMReasoning, NativeMultimodal & 3 \\
Biology & ProteinDesign, SingleCellFoundation, SpatialTranscriptomics & 3 \\
Chemistry & DrugDiscovery, MolecularGeneration, Retrosynthesis & 3 \\
Climate & AtmosphericChemistry, ClimateSimulation, WeatherForecasting & 3 \\
Energy & BatteryOptimization, PerovskiteSolarCell, SmartGridOptimization & 3 \\
Materials & BatteryElectrolyte, Catalysis, Discovery & 3 \\
Mathematics & AIforCombinatorics, AutomatedTheoremProving, MathReasoning & 3 \\
Medicine & DrugRepurposing, MedicalFoundationModels, MedicalImageAnalysis & 3 \\
Neuroscience & BrainComputer, Connectomics, VisualCortex & 3 \\
Physics & FusionPlasmaControl, NeuralNetworkPotentials, QuantumErrorCorrection & 3 \\
\midrule
\textbf{Total} & & \textbf{30 active} \\
\bottomrule
\end{tabular}
\end{table}

\section{\gene{} Examples}
\label{app:examples}

\subsection{\geneexam{}}
\label{app:ex_exam}

We present one representative example per capability dimension (T1--T4), showing the full question, choices, and gold answer.

\begin{tcolorbox}[
    breakable,
    title={Example of T1 · Genome Abstraction (T1-01: Contribution Role)},
    colback=LighterGray,
    colframe=DeepPurple,
    colbacktitle=DeepPurple,
    coltitle=White,
]
\textbf{\textcolor{CaseOrange}{Question.}} Given an anonymized genome extracted from a paper, classify its contribution type.
Options: \texttt{method}, \texttt{dataset}, \texttt{analysis}, \texttt{system}, \texttt{theory}.

\medskip
\textit{Claim}: ``A probabilistic model that accounts for repertoire-specific maturation age can reliably recover the true landscape and length statistics of antibody somatic indels from sequencing data.''\\
\textit{Delta}: ``Provides a dedicated probabilistic method for unbiased learning of antibody indel statistics, compared with prior work relying on biased annotation-based indel calls.''\\
\textit{Observation}: ``Applied to large human heavy-chain datasets, the model finds distinct insertion/deletion hotspots and shows indel lengths are approximately geometrically distributed.''

\medskip
\textbf{\textcolor{CaseBlue}{Answer:}} \texttt{method} \quad
\textit{(The paper introduces a new probabilistic model — a clear methodological contribution.)}
\end{tcolorbox}

\bigskip

\begin{tcolorbox}[
    breakable,
    title={Example of T2 · Inheritance Tracing (T2-04: Lineage Grouping)},
    colback=LighterGray,
    colframe=DeepPurple,
    colbacktitle=DeepPurple,
    coltitle=White,
]
\textbf{\textcolor{CaseOrange}{Question.}} Eight \genome{} objects are presented in shuffled order. Partition them into exactly two lineage groups of four objects each, ordered from earliest to latest.

\smallskip
\textbf{P}: \textit{``Directly optimize a neural architecture policy using RL to assign layer-wise operations.''}\\
\textbf{Q}: \textit{``NIN replaces fully-connected layers with global average pooling and $1{\times}1$ conv to deepen non-linearity.''}\\
\textbf{R}: \textit{``DARTS relaxes the discrete architecture search space into a continuous one via differentiable mixed-operation weights.''}\\
\textbf{S}: \textit{``Inception-v3 factorizes large convolutions into asymmetric sequences to reduce parameters.''}\\
\textbf{T}: \textit{``EfficientNet jointly scales width, depth, and resolution via compound coefficients.''}\\
\textbf{U}: \textit{``ENAS shares weights across child networks to reduce NAS search cost by orders of magnitude.''}\\
\textbf{V}: \textit{``GoogLeNet stacks Inception modules to achieve high accuracy at low parameter cost.''}\\
\textbf{W}: \textit{``MnasNet uses RL-based NAS with mobile hardware latency as a proxy reward.''}

\medskip
\textbf{\textcolor{CaseBlue}{Answer:}}
\begin{quote}\small
\textbf{Group 1 (NAS-RL lineage):} P $\to$ U $\to$ R $\to$ W\\
\textbf{Group 2 (Inception lineage):} Q $\to$ V $\to$ S $\to$ T
\end{quote}
\textit{Why}: Group 1 shares a lineage of learning architectural decisions (RL search $\to$ weight sharing $\to$ differentiable relaxation $\to$ mobile NAS); Group 2 shares the Inception niche of efficient multi-scale feature extraction.
\end{tcolorbox}

\bigskip

\begin{tcolorbox}[
    breakable,
    title={Example of T3 · Evolutionary Dynamics (T3-01: Single-Step Dynamics)},
    colback=LighterGray,
    colframe=DeepPurple,
    colbacktitle=DeepPurple,
    coltitle=White,
]
\textbf{\textcolor{CaseOrange}{Question.}} Given the Niche and Mechanism of predecessor A and successor B, identify the primary evolutionary dynamic.

\smallskip
\textbf{Predecessor A:}\\
\textit{Niche}: ``Understanding and improving robustness of neural network classifiers to small, intentionally constructed input perturbations that cause misclassification.''\\
\textit{Mechanism}: ``FGSM linearizes the loss at the input point, computes the gradient w.r.t.\ the input, takes its element-wise sign, and adds a small $\epsilon$-perturbation under $\ell_\infty$ constraint to generate an approximate worst-case sample.''

\smallskip
\textbf{Successor B:}\\
\textit{Niche}: ``Understanding \emph{why} deep neural network classifiers remain vulnerable to adversarial perturbations despite strong generalization on natural inputs.''\\
\textit{Mechanism}: ``Uses first-order linear analysis to explain adversarial examples, modeling classifiers as locally linear in high-dimensional input space: a perturbation chosen along the input gradient accumulates across many dimensions to significantly alter class scores.''

\medskip
\textbf{\textcolor{CaseOrange}{Choices:}}
\begin{quote}\small
A.\ Mutation \quad B.\ Adaptive Radiation \quad C.\ Hybridization \quad D.\ Speciation \quad E.\ Niche Competition
\end{quote}

\textbf{\textcolor{CaseBlue}{Answer: A.\ Mutation}}\\
\textit{Why}: B inherits and refines A's first-order gradient-sign mechanism, while narrowing to a theoretical niche (explaining vulnerability) — a local modification, not a structural departure.
\end{tcolorbox}

\bigskip

\begin{tcolorbox}[
    breakable,
    title={Example of T4 · Structural Integrity (T4-06: Missing-Link Recovery)},
    colback=LighterGray,
    colframe=DeepPurple,
    colbacktitle=DeepPurple,
    coltitle=White,
]
\textbf{\textcolor{CaseOrange}{Question.}} Paper A and Paper C are given; identify the missing bridge paper B from four candidates.

\smallskip
\textbf{Paper A}: ``Balance control for legged robots using ZMP-based model predictive control; focuses on stability under static and slow dynamic gaits.''

\textbf{Paper C}: ``Sim-to-real RL for agile locomotion; trains a neural network policy end-to-end in simulation and deploys on hardware using domain randomization.''

\medskip
\textbf{\textcolor{CaseOrange}{Candidates:}}\\
\textbf{X}: ``A dataset of human gait trajectories for imitation learning.''\\
\textbf{Y}: ``A periodic gait stabilization framework that decouples cycle generation from balance correction, enabling classical controllers to handle dynamic terrain.''\\
\textbf{Z}: ``A transformer-based language model for robot task planning.''\\
\textbf{W}: ``A convolutional terrain classifier for foothold selection.''

\medskip
\textbf{\textcolor{CaseBlue}{Answer: Y}}\\
\textit{Why}: Y bridges A's model-predictive stability focus and C's RL-based locomotion by introducing a periodic gait framework — a structural predecessor that made dynamic, policy-compatible gait generation feasible before the RL era.
\end{tcolorbox}

\subsection{\genearena{}}
\label{app:ex_arena}

\begin{tcolorbox}[
    breakable,
    title={Example of \genearena{} · Pairwise Idea Battle (Physics / Fusion Plasma Control)},
    colback=LighterGray,
    colframe=DeepPurple,
    colbacktitle=DeepPurple,
    coltitle=White,
]
\textbf{\textcolor{CaseOrange}{Question.}} Given a frontier research question and two anonymously generated ideas, judge which idea is stronger based on lineage grounding, meaningful variation, and selection value for future research. Options: \textbf{A}, \textbf{B}, or \textbf{Tie}.

\medskip
\textit{Frontier question}: Predict-first TCV and RL plasma-control studies show that data-driven control can learn device-specific dynamics, but ITER-scale deployment needs cross-device transfer, uncertainty-aware safety, and simulator-to-shot adaptation. What fusion-control foundation model would combine differentiable plasma simulators, multi-tokamak logs, and constrained RL to produce robust ramp-up, sustainment, and ramp-down policies across machines?

\medskip
\textbf{Idea A — Uncertainty-Aware Tokamak Control Foundation Model.}
Pretrain a transformer/state-space latent dynamics model on multi-tokamak time-series data (DIII-D, TCV, JET, EAST, C-Mod, HL-3 and TORAX rollouts), coupled to differentiable TORAX physics with learned residuals. Train constrained model-based RL policies with Bayesian uncertainty estimates and runtime safety shields (tearing, shape, density, disruption budgets). Before deployment, calibrate via a small number of predict-first target-machine shots updating Bayesian residuals — without retraining the full model.

\medskip
\textbf{Idea B — FusionFoundationRL: Cross-Tokamak Differentiable-Constrained Policy.}
Combine TORAX's differentiable physics engine with a neural state-space model (NSSM) into a hybrid dynamics foundation model trained on multi-tokamak logs. Device-specific parameters (geometry, coil limits) are encoded as context vectors. A constrained RL agent uses cross-machine tearing/disruption predictors as safety bounds, and TORAX's differentiability initializes the policy via gradient-based optimization. Unified ramp-up, sustainment, and ramp-down in a single policy, with simulator-to-shot fine-tuning for adaptation.

\medskip
\textit{Which idea better addresses the frontier? Consider: how well does each inherit and build on prior work, how meaningfully does it advance beyond it, and how much does it open new research directions?}
\end{tcolorbox}

\bigskip

\begin{tcolorbox}[
    breakable,
    title={Example of \genearena{} · Battle with Judge Result (Biology / Protein Design)},
    colback=LighterGray,
    colframe=DeepPurple,
    colbacktitle=DeepPurple,
    coltitle=White,
]
\textbf{\textcolor{CaseOrange}{Frontier question.}}
Design de novo protein binders against therapeutic targets where high in silico affinity does not reliably translate into selective, stable, functional activity in living cells.
What generative framework would close the loop between computational design and in vivo biological function?

\medskip
\textbf{Idea A — Cell-in-the-Loop Multi-Objective Protein Binder Design.}
Couple structure-based binder generation (diffusion/flow-matching conditioned on target epitope, off-target panels, and stability constraints) with a learned multi-fidelity fitness model. Candidate designs are scored by a differentiable surrogate ensemble covering binding affinity, target selectivity, folding stability, expression, and cell-based efficacy. A causal latent variable model separates true target engagement from confounders (expression level, trafficking, toxicity). The surrogate is updated iteratively via active learning from barcoded pooled yeast/mammalian display and cellular reporter assays, converting protein binder design into a self-improving closed-loop system optimized for experimentally meaningful activity.

\medskip
\textbf{Idea B — Multi-Constraint Latent Diffusion for In-Vivo-Aware Protein Binder Design.}
Learn a unified latent representation from multi-modal data (structures, binding assays, biophysical stability, in vivo functional readouts) using a multi-modal VAE, then run a conditional diffusion model in this latent space guided by target structure embeddings and a learned biological constraint manifold. A differentiable in vivo efficacy surrogate provides gradient feedback during generation; selectivity is enforced by contrastive learning against off-target proteomes; stability is modeled via physics-informed neural networks. A Pareto-frontier multi-objective reward balances competing constraints, aiming to raise experimental validation rates from $\sim$5--10\% to $>$30\%.

\medskip
\textbf{\textcolor{CaseBlue}{Judge result: Idea A wins.}}\\
\textbf{\textcolor{CaseBlue}{Judge reason:}} \textit{``Idea A more cleanly extends the frontier lineage by preserving structure-first generative design while adding assay-grounded active learning and causal components that directly target the known in silico-to-function gap, with a more feasible validation plan. Idea B is ambitious but relies on weakly transferable in vivo surrogates and expensive mouse-model validation.''}
\end{tcolorbox}

\section{Evaluation Details}
\label{app:eval_logging}

All \geneexam{} runs write the decoding budget into the top-level JSON metadata as \texttt{max\_output\_tokens}. Per-instance logs also store input/output token counts when returned by the API, response length, extraction diagnostics, and an error class. Empty responses are therefore separated into API failures, missing-format outputs, and token-budget exhaustion rather than being silently counted as ordinary content errors.

This detail matters for reasoning-heavy deployments. In pilot GPT-5.5 runs, a 4,096-token completion cap sometimes produced empty visible answers while the API reported the full output budget as consumed. We therefore use a 16,384-token cap for GPT-5.5 main-challenge reruns and keep the cap in the released logs so that scores remain auditable across providers and reruns. The paper leaderboard reports the same exact-match metric for all models; token-budget metadata is used only for diagnosis and reproducibility, not for changing an instance score.

\section{Human Agreement}
We recruited 50 graduate annotators (master’s and doctoral students across
computer science, biology, physics, materials science, and other disciplines) to validate the benchmark
at three stages: \genomediff{} relation labels and role-type assignments during construction (disagreements adjudicated; ambiguous items excluded), \geneexam{} item difficulty through stratified
human solving to confirm low accuracy reflects compositional challenge rather than label noise, and
\genearena{} ELO battles where human judges reached 80\% agreement with the strongest model-judge panel on pairwise preference comparisons. PES scoring uses the same lineage packet format for model and human audit: judges see the frontier question, relevant \genome{} objects, \genomediff{} evidence, and population neighborhood before assigning H/V/S. This concordance supports scalable model-judge evaluation while confirming that scoring criteria are reproducible by trained domain experts.

\section{Additional \genearena{} Analysis}
\label{app:arena_analysis}

\subsection{Domain-Level PES}

This view checks whether lineage-grounded generation quality is uniform across scientific domains. Each cell averages Lineage-setting PES over the three active frontier tasks in that domain, so row-wise variation exposes domain-specific weaknesses that are hidden by a single overall PES.

\begin{figure}[h]
\centering
\includegraphics[width=\textwidth]{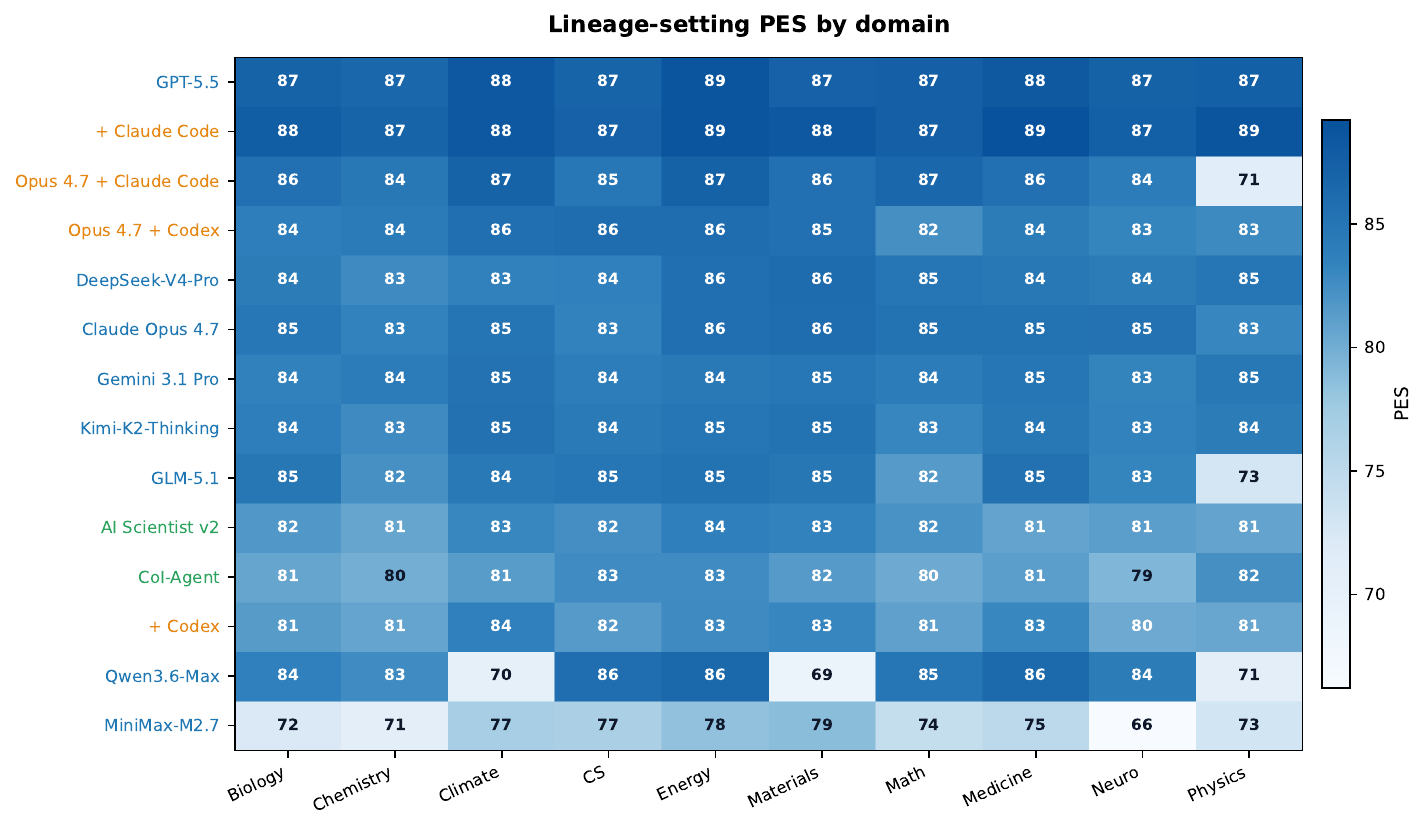}
\caption{\textbf{Domain-level PES snapshot.} Lineage-setting PES across 10 domains for the evaluated systems. Strong systems remain relatively uniform, while weaker systems show sharper domain-specific failures.}
\label{fig:pes_domain_app}
\end{figure}

\subsection{PES Decomposition and Information-Setting Breakdown}

This pair of heatmaps decomposes PES along two orthogonal axes. Panel~(a) separates the three rubric components---Heredity, Variation, Selection---to show whether high PES comes from coherent inheritance, meaningful novelty, or downstream viability; the recurring Variation-over-Heredity pattern is the main evidence for the plausibility--coherence gap discussed in the main text. Panel~(b) isolates what each information setting contributes: Question-only prompts test parametric ideation, Library adds unordered paper context, and Lineage adds ordered \genome{} objects plus \genomediff{} evidence; heterogeneous gains show that lineage evidence is useful only when a system can operationalize it.

\subsection{Generated Dynamics Distribution}

This figure preserves the raw frequency view replaced by the main-text dynamics quality map. It shows that Question-only prompts produce a broader mix of evolutionary moves, while Library and Lineage prompts concentrate on Hybridization; by itself, however, this distribution cannot tell whether the recombination is genetically coherent.

\subsection{Per-System H/V/S Bar Chart}

This bar chart provides a ranking-oriented view of the same Lineage-setting H/V/S decomposition shown in the heatmap above. Systems are sorted by Lineage PES; the visual separation between Variation and Heredity bars confirms that the plausibility--coherence gap is systematic across all 14 participants.

\begin{figure}[h]
\centering
\includegraphics[width=0.58\textwidth]{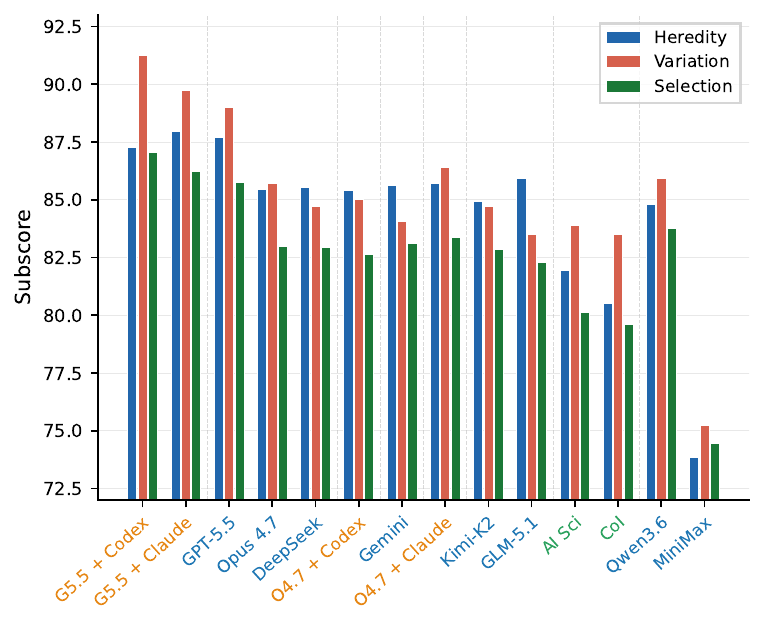}
\caption{\textbf{Per-system H/V/S bar chart.} Lineage-setting Heredity, Variation, and Selection for all evaluated systems, sorted by aggregate Lineage PES. Variation consistently exceeds Heredity, confirming the plausibility--coherence gap across the full participant set.}
\label{fig:pes_bars_app}
\end{figure}

\FloatBarrier

\section{Additional \geneexam{} Analysis}
\label{app:exam_analysis}

\subsection{\geneexam{} Main-Leaderboard Heatmap}

This heatmap reports the main leaderboard's exact accuracy across T1--T4. We keep it here as the detailed closed-form diagnostic view while the main text uses the more compact radar profile.

\subsection{Error Analysis: Failure Flow}

This Sankey view aggregates wrong answers by capability tier, failed field family, and error class. It is meant to localize the bottleneck: most failures originate in evolutionary reasoning and verification, then propagate through driver and dynamics fields before becoming exact-match errors.

\begin{figure}[h]
\centering
\includegraphics[width=0.85\textwidth]{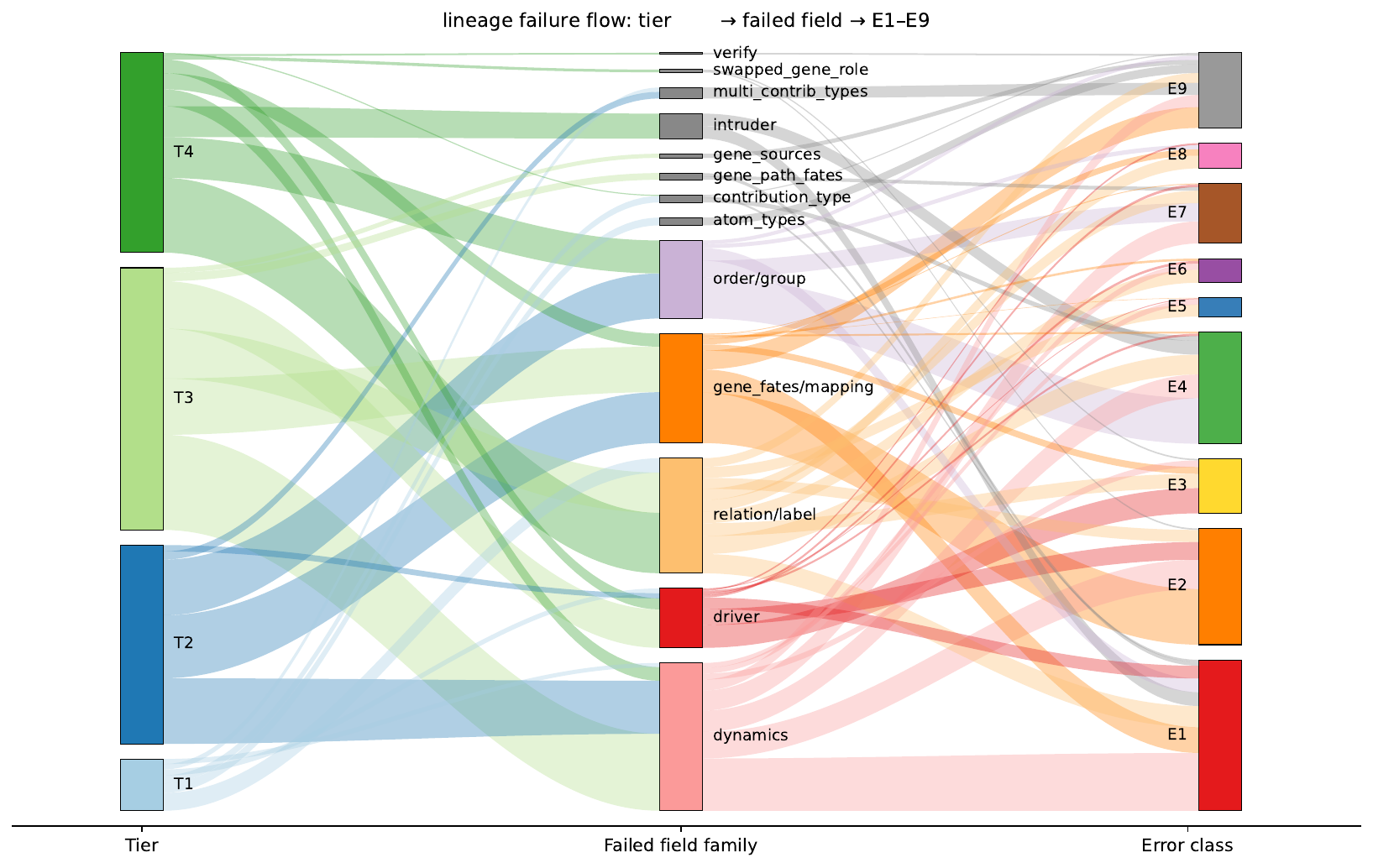}
\caption{\textbf{\geneexam{} failure flow.} Sankey diagram tracing errors from capability tier (T1--T4) through failed field family to error class (E1--E9). T3 and T4 errors dominate in volume. The thickest flows connect T3/T4 to dynamics and driver fields, which then fan out to E1 (dynamics misclassification) and E2 (driver misidentification). Genome-fate and relation errors (E3--E4) are the next largest classes, confirming that compositional field interactions---not isolated label failures---drive the accuracy bottleneck.}
\label{fig:error_sankey_app}
\end{figure}

\end{document}